\documentclass[10pt,twocolumn,letterpaper]{article}

\usepackage{cvpr}
\usepackage{times}
\usepackage{epsfig}
\usepackage{graphicx}
\usepackage{amsmath}
\usepackage{amssymb}
\usepackage{subfigure}

\usepackage[pagebackref=true,bookmarks=false]{hyperref}

\cvprfinalcopy 

\def\cvprPaperID{0014} 

\ifcvprfinal\fi
\begin{document}

\title{Learning Spatiotemporal Features for Infrared Action
  Recognition with 3D Convolutional Neural Networks}


\author{Zhuolin Jiang, Viktor Rozgic, Sancar Adali \\
Raytheon BBN Technologies, Cambridge, MA, 02138 \\
{\tt\small \{zjiang, vrozgic, sadali\}@bbn.com}
}

\maketitle

\begin{abstract}
Infrared (IR) imaging has the potential to enable  more robust action recognition systems compared to visible spectrum cameras due to lower sensitivity to lighting conditions and appearance variability. While the action recognition task on videos collected from visible spectrum imaging has received much attention, action recognition in IR videos is significantly less explored. Our objective is to exploit imaging data in this modality for the  action recognition task.

In this work, we propose a novel two-stream 3D convolutional neural network (CNN) architecture by introducing the \emph{discriminative code layer} and the corresponding discriminative code loss function. The proposed network processes IR image and the IR-based optical flow field sequences. We pretrain the 3D CNN model on the visible spectrum Sports-1M action dataset and finetune it on the Infrared Action Recognition (InfAR) dataset. To our best knowledge, this is the first application of the 3D CNN to action recognition in the IR domain. We conduct an elaborate analysis of different fusion schemes (weighted average, single and double-layer neural nets) applied to different 3D CNN outputs. Experimental results demonstrate that our approach can achieve state-of-the-art average precision (AP) performances on the InfAR dataset: (1) the proposed two-stream 3D CNN achieves the best reported  $77.5\%$ AP, and (2) our 3D CNN model applied to the optical flow fields achieves the best reported single stream $75.42\%$ AP.
\end{abstract}

\vspace{-0.5cm}
\section{Introduction}
\vspace{-0.1cm}

Deep convolutional neural networks (CNN) have shown remarkable
success for various computer vision tasks in static
images, such as object detection~\cite{rcnn}, recognition~\cite{hinton12} and segmentation~\cite{Long15}.  Encouraged by this success, researchers  have proposed some CNN-based algorithms  for action recognition in visible spectrum videos~\cite{shengxin15,Lan15,Donahue15,twostream1,twostream3,Joe15,feichtenhofer16,jiang17}. One promising approach is using a two-stream CNN architecture developed by~\cite{twostream1}, which consists of a spatial stream network for
learning salient appearance features from video frames, and a temporal stream network for
learning motion patterns. The prediction is computed by averaging the outputs of two networks. This architecture showed improved performance over traditional action recognition approaches such as improved dense trajectories features~\cite{idt}. However, as pointed out by~\cite{Tran15}, 2D convolutions in a temporal network applied on the multi-frame stacking of optical flow fields (treating them as different channels) generate 2D representations; and the temporal network loses temporal information important for action recognition after the first convolution layer. To address this, ~\cite{Tran15}
introduced a 3D CNN which takes multiple RGB frames as inputs and performs 3D convolution and pooling, preserving temporal information. The 3D CNN models can  process appearance and motion information simultaneously, hence it is able to learn spatiotemporal features for action recognition.

For action recognition in infrared videos, there is
limited work that uses deep CNNs to combine spatial and
temporal cues for spatiotemporal feature learning~\cite{infar,Zhu13}.~\cite{infar} applied
two-stream CNNs for infrared action recognition.
However, the CNN stream that processes the infrared image sequence achieved worse performance than several
hand-crafted low-level features such as spatio-temporal interest point~\cite{Laptev05}
and dense SIFT~\cite{sift}. There are two potential reasons for this: 1) the infrared InfAR dataset is not large enough to
learn spatiotemporal features leading to severe
overfitting; and 2) 2D CNN loses temporal information contained in an input video volume as pointed in~\cite{Tran15}, so it can not properly model the temporal action patterns.

In this paper, we propose a two-stream 3D CNN architecture to learn spatio-temporal features for infrared action recognition. The two-stream 3D CNN contains two separate recognition networks (IR and optical-flow nets) combined using late fusion. In order to reduce the chance of
overfitting and learn discriminative spatio-temporal
features, we incorporate a discriminative code loss
introduced in~\cite{jiang17}, and combine it with softmax
classification loss to form the objective function used for network training. For faster convergence
during training, we pretrained 3D CNN model parameters on
the large-scale Sports-1M action dataset~\cite{sports1M} with videos from the  visible light spectrum and finetuned
them on the infrared dataset. The results are surprisingly good.

Our main contributions are the following:
\begin{itemize}
\item We develop a two-stream 3D CNN architecture to learn spatiotemporal features from infrared videos. This two-stream model learns representations that capture spatial and temporal information simultaneously.
\vspace{-0.3cm}
\item We add a discriminative code layer (DCL) on top of the last fully-connected layer and combine the discriminative code loss with softmax classification loss to train the 3D CNN. This discriminative code layer generates class-specific representations for infrared videos.
\vspace{-0.1cm}
\item We achieve state-of-the-art performances on the InfAR
  dataset. We find that a single 3D CNN with DCL layer trained using the optical flow field images can achieve an excellent infrared action recognition performance.
\end{itemize}
\vspace{-0.2cm}

\subsection{Related work}
Many popular video feature extraction and classification approaches have been developed for action recognition in the visible spectrum, including low-level features (e.g., spatio-temporal interest point (STIP)~\cite{Laptev05}, scale-invariance feature transform (SIFT)~\cite{sift}, optical flow fields~\cite{lin09}, improved dense trajectory feature (IDT)~\cite{idt}), and high-level semantic concepts (e.g., human action attributes~\cite{liu11,Qiu11} and action parts~\cite{Yao11}). In recent years, approaches based on convolutional neural networks have been
proposed for action recognition~\cite{Lan15,Donahue15,twostream1,twostream3,Joe15,feichtenhofer16,LWang16}. The two-steam CNN architecture~\cite{twostream1} achieved impressive recognition performances.~\cite{twostream3} explored deeper two-stream network architectures and refined technical details yielding further performance improvements.~\cite{feichtenhofer16} investigated an effective fusion strategy over space and time for two steam networks. However, all state-of-the-art approaches focus on the action recognition in the visible light spectrum.

Compared with visible spectrum imaging, the infrared imaging has a nice property that it can work well even under poor light conditions, which is useful for nighttime surveillance~\cite{Zhu13,infar,Tran15}. Still, limited works address the action recognition in infrared spectrum.~\cite{Zhu13} trained a SVM classifier trained on visible light spectrum based bag-of-visual-words video representation and adapted it to the infrared domain.
Recently, ~\cite{infar} applied a two-stream 2D CNN architecture to infrared videos. This architecture employs a motion-history-image (MHI) stream network and an optical-flow stream network to extract image-level features. When the MHI stream is replaced with the raw infrared image stream, the action recognition performance becomes very poor, showing that the two-stream 2D CNN highly relies on the MHI stream network.

Different to~\cite{infar}, we introduce a two-stream 3D
CNNs, which can model the video appearance and motion information simultaneously from an infrared video. To our best knowledge, 3D CNN architectures have not been explored for action recognition in infrared videos. In addition, we
integrated the discriminative code loss from~\cite{jiang17}
to the objective function of network training. This makes
the learned representations more discriminative and good for classification tasks, specifically action recognition. These two sources of novelties enable an action recognition system that advances state-of-the art performance for action recognition in IR videos, as evidenced in our experiments.

\section{Learning Spatiotemporal Features with 3D Convolutional Neural Networks}

\begin{figure}[t]
	\begin{center}
		\includegraphics[width=1.0\linewidth]{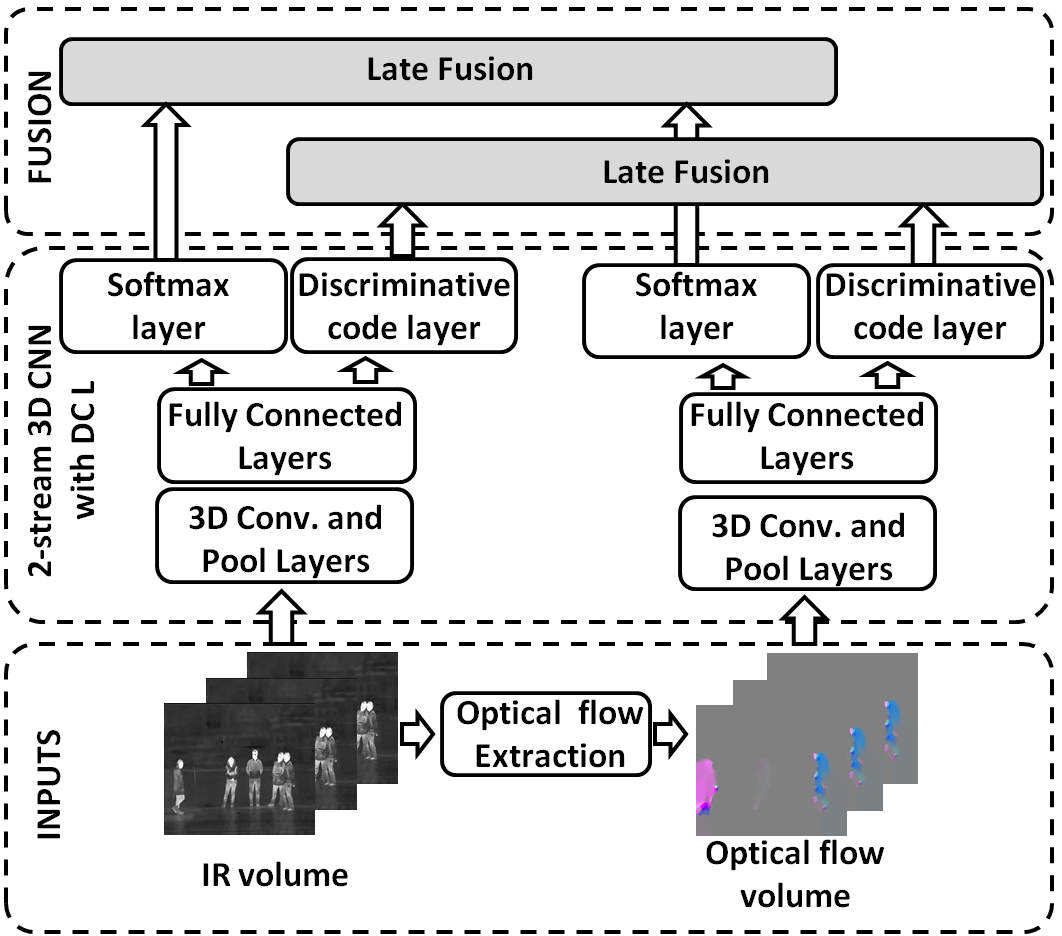}
	\end{center}
	\vspace{-0.1cm}
	\caption{The IR action recognition pipeline: a two-stream 3D CNN trained with softmax classification loss and discriminative code loss, and their outputs aggregated using late fusion.}
	\label{pipeline}
	\vspace{-0.3cm}
\end{figure}

Our pipeline for action recognition in IR videos is presented in Figure \ref{pipeline}. The pipeline inputs are IR video clips obtained by splitting IR frame sequences into non-overlapping segments of
consecutive frames. Motivated by the success of the two-stream CNN architectures operating on visible spectrum videos and derived optical flow fields, we convert each consecutive pair of IR frames into optical flow field.
We resize the IR and the optical flow images to the same height and width before feeding the IR and the optical flow clips to their corresponding networks. In the InfAR dataset experiments, we used a frame size of
$128\times171$, creating the input IR and optical flow clips with the same dimension $3\times t\times 128\times 171$, where $3$ is the number of channels of an IR or flow image~\footnote{We compute the optical flow
image according to \cite{Brox04}. We stack the $x$, $y$ components of a flow field, and compute its magnitude as the third channel of the flow image. The flow values of $x$, $y$ components are centered around $128$
and scaled such that flow values fall between $0$ and $255$.} and $t$ is the temporal clip length in number of frames.

We process the corresponding IR and optical flow clips using a novel two-stream 3D CNN architecture. Both streams are processed using the same CNN model which extends the existing 3D-CNN architecture~\cite{Tran15}
by introduction of the additional \emph{discriminative code layer}. The \emph{discriminative code loss} associated with the discriminative code layer is combined with the softmax classification loss to train the 3D CNN.

We fuse the probabilistic outputs from the softmax (or the discriminative code) layers of the proposed two-stream network by using the weighted average, the single-layer neural network (NN) fusion or the two-layer NN fusion.

\subsection{3D Convolutional Neural Network with Discriminative Code Layer}
Compared to 2D CNN architecture, 3D CNN architecture is better suited for the action recognition task, because it models spatial and temporal information jointly using 3D convolution and 3D pooling operations.
While 2D convolution architectures process multiple input frames as different input channels and transform them into 2D representations; 3D convolution transforms input volume into 3D representation preserving temporal
information. To our best knowledge, 3D CNNs have not yet been applied to the task of action recognition in IR videos.

\begin{figure*}[t]
	\begin{center}
		\includegraphics[width=0.7\linewidth]{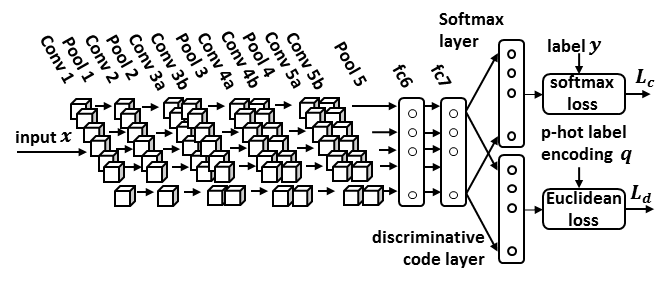}
	\end{center}
	\vspace{-0.1cm}
	\caption{3D CNN is trained using the cost function that combines the softmax classification loss and the discriminative code loss.}
	\label{3DCNN}
	\vspace{-0.2cm}
\end{figure*}

Our two-stream 3D CNN architecture is based on the 3D CNN model with the \emph{discriminative code layer} presented in Figure \ref{3DCNN}. The 3D CNN follows the architecture proposed in \cite{Tran15}.
The network has eight 3D convolution layers, combined using five 3D max-pooling layers. The last pooling layer is followed by two fully connected layers (fc6 and fc7). The details on sizes of convolutional kernels,
numbers of filters in different convolutional layers, sizes of the max-pooling and fully connected layers are provided in Section \ref{sec:exp:setting}.

The proposed 3D CNN architecture has two output layers, a \emph{softmax layer} that generates $m$-dimensional one-hot encoding of $m$ activity categories and a \emph{discriminative code layer} that generates discriminative
codes for inputs signals. Assuming the discriminative code layer with $N$ neurons, the training goal is to make it generate $N$ dimensional $p$-hot encoding~\footnote{here we assume that $N$ can be exactly divided by $m$,
~\textit{i.e.,} $N=pm$. If $N=pm+k (k<m)$ we allocate the remaining $k$ neurons to classes with high intra-class appearance variations.} of activity categories. The $p$-hot encoding represents a sample from $k^{th}$ action
category ($k=1,...,m$) as a binary vector with coordinates $[p(k-1)+1:pk]$ equal to one and rest of coordinates equal to zero. Intuitively, the group of neurons activates only when a sample from the corresponding category is
presented. In order to achieve this, we introduce the \emph{discriminative code loss} associated with the discriminative code layer activations. This loss encourages groups of output neurons to activate simultaneously
encoding the category label.

Let's assume that the 3D CNN architecture has $n+1$ layers, $n$ levels including all convolution layers, pooling layers and fully connected layers, and the layer $n+1$ including the softmax layer and discriminative
code outputs. The output of the $i^{\text{th}}$ layer is denoted as $\mathbf{x}^{(i)}$, where $\mathbf{x}^{(0)}$ represents input. Therefore, the network architecture can be concisely expressed as:
\begin{align}
&\mathbf{x}^{(i)} = F(\mathbf{W}^{(i)}\mathbf{x}^{(i - 1)}),\quad i = 1, 2, ..., n \label{eq1} \\
&\mathbf{x}_d^{(n+1)} = \mathbf{A}\mathbf{x}^{(n)}                                 \label{eq2} \\
&\mathbf{x}_c^{(n+1)} = \mathrm{softmax}(\mathbf{W}\mathbf{x}^{(n)})               \label{eq3}
\end{align}
where $\mathbf{W}^{(i)}$ represents the network parameters\footnote{For simplicity, we ignore the bias term for convolution layers, fully connected layers, softmax layer and discriminative code layer.\label{footnote3}} of the $i^{\text{th}}$ layer for convolution and fully-connected layers, $\mathbf{W}^{(i)}\mathbf{x}^{(i - 1)}$ is a linear operation (\textit{e.g.}
convolution in a convolution layer, or a linear transformation in fully-connected layer), $F(\cdotp)$ is a non-linear activation function (\textit{e.g.} ReLU). $\mathbf{A}$ contains parameters of a linear transformation
implemented by the \emph{discriminative code layer} and $\mathbf{W}$ contains the \emph{softmax layer} parameters. $\mathbf{x}_d^{(n+1)}$ is the \emph{predicted code} while $\mathbf{x}_c^{(n+1)}$ is the predicted class score
vector.

The overall loss function used in network training is a linear combination of the softmax classification loss (multinomial logistic loss) $L_c$, and \emph{discriminative code loss} $L_d$ with a cost-balancing hyper parameter $\alpha$.
\begin{align}
& L = L_c + \alpha L_d                                                    \label{eq3}\\
& L_c = L_c(\mathbf{x}_c^{(n+1)}, y)                              \label{eq4}\\
& L_d = L_d(\mathbf{x}_d^{(n+1)}, y)                              \label{eq5}
\end{align}

The $L_d$ cost component can be defined as:
\begin{equation} \label{eq6}
L_d = \lVert \mathbf{q}^{(n)} - \mathbf{A}\mathbf{x}^{(n)} \rVert_2^2,
\end{equation}
where the binary vector $\mathbf{q}^{(n)} = [q^{(n)}_{1}, \ldots,q^{(n)}_{j},\ldots, q^{(n)}_{N}]^{\rm{T}} \in \{0, 1\}^{N}$ denotes the $p$-hot label encoding (or target discriminative code), which indicates the ideal
activations of neurons ($j$ denotes the index of neuron. Each neuron is associated with a certain class label and, ideally, only activates to samples from that class. Therefore, when a sample is from the $k^{th}$ action
category, $q^{(n)}_{j} = 1$ if and only if the $j^{\rm{th}}$ neuron is assigned to class $k$, and neurons associated to other classes should not be activated so that the corresponding entries in $\mathbf{q}^{(n)}$ are zero.
Note that $\mathbf{A}$ is the only parameter\textsuperscript{\ref{footnote3}} to be learned in this cost component.

\subsection{Network Training}

The network parameters $(\mathbf{W}^{(i)})_{i = 1, 2, \ldots, n}),\emph{A},\emph{W})$ are trained via back-propagation using the mini-batch stochastic gradient descent method. Compared to the parameter update equations for a multi-layer CNN~\cite{LeCun12} without the discriminative cost loss, the gradient term, \textit{i.e.} $\frac {\partial L} {\partial \mathbf{x}^{(n)}}$ changes and two gradient terms $\frac {\partial L} {\partial \mathbf{A}}$ and $\frac {\partial L} {\partial \mathbf{x}_d^{(n+1)}}$ are introduced, since $\mathbf{x}^{(n)}$, $\mathbf{A}$ and $\mathbf{x}_d^{(n+1)}$ are related to the discriminative code loss $L_d$.

From Equations (\ref{eq4}) and (\ref{eq5}), we can calculate $\frac {\partial L} {\partial \mathbf{x}_d^{(n+1)}}$ and $\frac {\partial L} {\partial \mathbf{x}_c^{(n+1)}}$. Then we can obtain $\frac {\partial L} {\partial \mathbf{A}}$, $\frac {\partial L} {\partial \mathbf{W}}$ and $\frac {\partial L} {\partial \mathbf{x}^{(n)}}$ by applying the chain rule:

\begin{align}
\frac {\partial L} {\partial \mathbf{x}_d^{(n+1)}} &= \alpha \frac {\partial L_d} {\partial \mathbf{x}_d^{(n+1)}}, \quad \frac {\partial L} {\partial \mathbf{x}_c^{(n+1)}} = \frac {\partial L_c} {\partial \mathbf{x}_c^{(n+1)}} \\
\frac {\partial L} {\partial \mathbf{A}} &= 2\alpha(\mathbf{A}\mathbf{x}^{(n)} - \mathbf{q}^{(n)})\mathbf{x}^{(n){\rm T}},  \frac {\partial L} {\partial \mathbf{W}} = \frac {\partial L_c} {\partial \mathbf{W}} \\
\frac {\partial L} {\partial \mathbf{x}^{(n)}} &= \frac {\partial L} {\partial \mathbf{x}_c^{(n+1)}}\frac {\partial \mathbf{x}_c^{(n+1)}}{\partial \mathbf{x}^{(n)}} + 2\alpha(\mathbf{A} \mathbf{x}^{(n)} - \mathbf{q}^{(n)})^{\rm T} \mathbf{A}
\end{align}

Once the partial derivative of $L$ with respect to $\mathbf{x}^{(n)}$ is known, the partial derivative of $L$ with respect to $\mathbf{W}^{(i)}$ and $\mathbf{x}^{(i-1)}$ can be computed using the backward recurrence:
\begin{align}
\frac {\partial L} {\partial \mathbf{W}^{(i)}} &= \frac {\partial L}{\partial \mathbf{x}^{(i)}}\frac {\partial \mathbf{x}^{(i)}}{\partial \mathbf{W}^{(i)}},  \nonumber \\
\frac {\partial L} {\partial \mathbf{x}^{(i-1)}} &= \frac {\partial L}{\partial \mathbf{x}^{(i)}}\frac {\partial \mathbf{x}^{(i)}}{\partial \mathbf{x}^{(i-1)}}, \quad \forall i \in \{1,...,n\}
\end{align}
where $\frac {\partial \mathbf{x}^{(i)}}{\partial \mathbf{W}^{(i)}}$ and $\frac {\partial \mathbf{x}^{(i)}}{\partial \mathbf{x}^{(i-1)}}$ can be computed from Equation (\ref{eq1}).


\subsection{Fusion}
As a part of the two-stream pipeline, we perform fusion of the probabilistic outputs from the IR and the optic flow nets. While the softmax layer directly provides a probabilistic output, we propose a method to convert the discriminative code layer outputs
to a multinomial distribution over action classes. Given the predicted code of a test sample, we find its $k$ nearest neighbors from each class in the training set and calculate the average distances from the sample to its $k$ neighbor training samples from each class. Then, we convert a set of average distances to sample-to-class similarity weights using a Gaussian kernel. Finally, we obtain a probability vector over action categories by ${\ell}_1$ normalization of the similarity weights.

We fuse the probability outputs of the IR and the optical flow nets using a simple weighted average approach. In addition, we apply two neural network based methods to fuse the predicted codes from IR and optical flow nets: (1) we concatenate predicted codes from from the IR and flow nets and use the obtained vector as an input to a single softmax layer neural network, which outputs probability estimates for action classes (\textbf{single-layer NN fusion}); and (2) we use the concatenated predicted codes as inputs to the two-layer neural network consisting of one $1\times 1$ convolution layer and the softmax output layer (\textbf{two-layer NN fusion}).

\section{Experiments}
\label{sec:exp}

\begin{figure*}
\begin{center}
\includegraphics[width=0.8\linewidth]{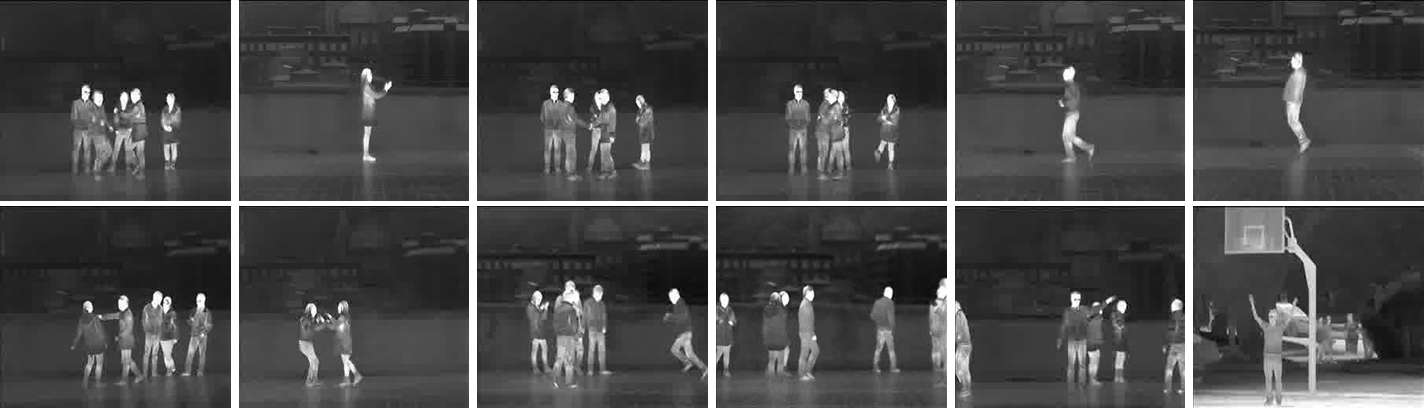}
\end{center}
\caption{Video samples for $12$ action
  classes from the InfAR action dataset~\cite{infar}. The action
  classes are `fight', `handclap', `handshake', `hug', `jog', `jump', `punch', `push', `skip', `walk',
  `wave1' and `wave2'.}
\label{fig3-1}
\vspace{-0.3cm}
\end{figure*}

We evaluate our approach on the recently released InfAR video
dataset~\cite{infar}, which is collected using infrared cameras. This dataset contains videos  of
$12$ different action classes~\footnote{The action
  classes include: `0-fight', `1-handclap', `2-handshake',
  `3-hug', `4-jog', `5-jump', `6-punch', `7-push', `8-skip', `9-walk',
  `10-wave1' and `11-wave2'.} with $50$ videos in each class. Figure~\ref{fig3-1} shows video
examples from the dataset. First, using this
dataset, we evaluate  the performance of  three widely used low-level descriptor features for the action class prediction task :  dense SIFT (D-SIFT)~\cite{Boureau10}, opponent
SIFT (O-SIFT)~\cite{Sande10}, and
motion features - improved dense trajectories features
(IDT)~\cite{idt}. Then, we evaluate the prediction performance of semantic feature vector produced  by $2,784$ concept detectors, each of which outputs a concept score, given the  low-level feature vector (e.g. D-SIFT) . Finally, we evaluate the two-stream 3D CNNs with the discriminative code layer using different output fusion strategies.

\subsection{Experimental Settings}
\label{sec:exp:setting}

To extract low-level image-based features such as dense SIFT
and opponent SIFT, we uniformly sample $50$
frames per video. We use Fisher vector (FV) encoding~\cite{Perronnin10} to
obtain video-level representations from these local low-level descriptors.  The video-level features are computed by spatio-temporal pooling of the frame-based FV features. In addition to these low-level video representations,  we extract features whose dimensions correspond to measures of evidence of high-level concepts in videos. For extracting semantic concept features, we trained $2,784$
concept detectors utilizing the VideoStory
Dataset~\cite{Habibian14}. The detectors are trained to predict high-level concept features from  different FV-encoded features
(D-SIFT, O-SIFT and IDT). Finally, we trained multiple linear multi-class SVM classifiers to predict actions from different low-level features and concept features.

The architectures of IR net and Flow net are identical and follow~\cite{Tran15}. Assume $C(k,n,s)$ is a convolutional layer with kernel size $k\times k \times k$, $n$ filters and stride $s\times s\times s$,
$P(k1,k2,s1,s2)$ is a max-pooling layer with kernel temporal size $k1$, kernel spatial size $k2\times k2$, temporal stride $s1$ and spatial stride $s2$. $FC(n)$ is a fully connected layer with $n$ filters.
$SM(n)$ is a softmax layer with $n$ filters. $DC(n)$ is a discriminative code layer with $n$ filters. The main architecture follows: $C(3,64,1)$--$P(1,2,1,2)$--$C(3,128,1)$--$P(2,2,2,2)$--$C(3,256,1)$--$C(3,256,1)$--$P(2,2,2,2)$--$C(3,512,1)$--$C(3,512,1)$--$P(2,2,2,2)$--$C(3,512,1)$--$C(3,512,1)$--$P(2,2,2,2)$--$FC(4096)$--$FC(4096)$--$SM(12)$($DC(4096)$).

The temporal length $t$ of each IR and flow clip is $16$ frames. The parameter $\alpha$ in Eq~\ref{eq3} is set to be $0.02$ in our
experiments. Both IR stream and Flow stream nets are pretrained on the
large-scale Sports-1M dataset~\cite{sports1M} and
finetuned on the InfAR dataset. The learning rate,
training batch size, weight decay coefficient and maximum
iterations are set as $0.0001$, $30$, $0.0005$ and $10,000$,
respectively. To extract video-level CNN features, we split a video into
$16$ frame long clips without overlapping between two
consecutive clips. To get the video-level representations, we simply averaged
representations extracted from each clip belonging to the video. To
classify actions, we use two ways: (1) employ the softmax output
layer to produce the confidence scores for all action classes for each video clip and use their average to predict the video-level class label (\textbf{softmax}); (2) employ the
$k$-NN classification based on the video-level representation, which is computed as the average of predicted codes of all video clips belonging to the video (\textbf{$k$-NN}).

We follow the standard setting in~\cite{infar}, we randomly select $30$ samples from each category as training, and the rest for testing. We repeat the experiments five times and report their performance average as the final performance in this paper. For the evaluation metrics, we used average precision (AP) as in~\cite{infar}, which is the average of recognition precisions of all actions.

\subsection{Comparisons with Other Approaches}

\subsubsection{Low-level and high-level semantic features}
\label{sec:exp:lowlevel}

We evaluate two static appearance features (D-SIFT and
O-SIFT), one motion feature (IDT) and the corresponding high-level
semantic concept features, all extracted from infrared
videos. We perform early SVM fusion by concatenating concept feature vectors
obtained using concept detectors on different low-level features. Finally, we perform late fusion by averaging the posterior scores of SVMs
trained on different features.

Table~\ref{tb1} summarizes the recognition performances with
these approaches. The high-level concept
features achieved similar or better performance compared
with the corresponding low-level features. In addition, the early
fusion of all concept features provided similar results as
the late fusion approach that combined the prediction scores from
six SVM classifiers.


\begin{table}
\centering
\begin{tabular} {|c|c|}
\hline
Method & AP (\%)\\
\hline
D-SIFT~\cite{Boureau10} & 46.7\\
D-SIFT based concepts & 46.7\\
O-SIFT~\cite{Sande10} & 47.5\\
O-SIFT based concepts & 47.1\\
IDT~\cite{idt} & 43.3 \\
IDT based concepts & 44.6 \\
Early fusion of all concepts & 47.5 \\
Late fusion of all features & 47.9 \\
\hline
\end{tabular}
\vspace{5pt}
\caption{Recognition performance comparisons in terms of average precisions (\%) using low level
  features and their corresponding semantic concept features.}
\label{tb1}
\end{table}

\subsubsection{Discriminative features from single 3D CNN}

Our two stream 3D-CNN is based on the C3D architecture
in~\cite{Tran15}. It consists of an IR net taking 16-frame clips
as inputs and a flow net taking 16-frame sequences of the optical flow fields.
We trained a 3D CNN in two ways:
(1) using softmax classification loss only; (2) using both
softmax classification loss and the discriminative code loss (DCL).

We first trained IR and Flow nets using softmax
classification only. We called these two networks as `IR
net without DCL' and `Flow net without DCL',
respectively. Then we maintain the softmax layer in both
networks but add the
discriminative code layer to the $fc_7$ layer to train two
stream networks, which are referred to as `IR net' and
`Flow net'. For IR and flow nets, we can
employ softmax or $k$-NN classification methods. The performances of different training and classification
methods are presented in Table~\ref{tb2}.

\begin{table}
\centering
\begin{tabular} {|c|c|}
\hline
Method & AP (\%)\\
\hline
IR net without DCL & 48.75\\
IR net (softmax)& 52.91\\
IR net ($k$-NN)& 54.58\\
Flow net without DCL & 69.58\\
Flow net (softmax) & 72.91\\
Flow net ($k$-NN) & 75.42\\
\hline
Two-stream-CNN-1~\cite{infar} & 32.08 \\
Two-stream-CNN-2~\cite{infar} & 76.66 \\
\hline
\end{tabular}
\vspace{5pt}
\caption{Recognition results of 3D-CNNs trained with or
  without discriminative code loss, and using different classification methods. The results of
  `Two-stream CNN-1' and `Two-stream CNN-2' are copied from the
  original paper~\cite{infar}. }
\label{tb2}
\vspace{-0.3cm}
\end{table}

As shown in Table~\ref{tb1} and Table~\ref{tb2}, the `IR net without DCL' can
still obtain marginally better results than the late fusion of all
low-level features and their concept features. The `Flow
net without DCL' can obtain around $20$\% improvement on
average precision compared to its partner stream, which
demonstrate that the motion information is very important for action recognition.
The `$k$-NN' classification method achieved better performance
than the `softmax' method, due to the increase of inter-class distances in the
generated discriminative code space.

We compare our results with two-stream CNN results reported
in~\cite{infar}. `Two-stream CNN-1' uses a two-stream 2D
convolutional architecture~\cite{twostream1} consisting of an IR stream and
a flow stream as ours. `Two-stream CNN-2' is similar to `Two-stream CNN-1',
but the IR stream is replaced with a stream taking optical flow motion history images as inputs.
As pointed out in~\cite{Tran15}, 3D CNN can model
appearance and motion simultaneously, hence our IR net outperforms the two-stream-CNN-1, which is trained with IR images and
optical flow fields. The two-stream-CNN-2 achieves good results due to the use of motion
history images. However, the result of our flow net, which
is based on optical flow field only, is still comparable to
two-stream-CNN-2.

\subsubsection{Fusion of two 3D CNNs}

We investigate different fusion strategies for the outputs from
IR and flow nets:

\emph{Late fusion1.}
We fuse the softmax probability
outputs of IR and Flow nets using simple weighted average rule, where
the weight is $2$ for flow net and $1$ for IR net.

\emph{Late fusion2.}
Given a predicted code vector from the discriminative code layer, we compute the average
distances to each class $i$ based on its $k$ neighbor training samples from each class, and convert them
to a similarity vector using a Gaussian kernel function : $\exp(-\gamma d^2_i)$, where $d_i$ is the average distance to class $i$ and $\gamma$ is a normalization factor. Finally, we
compute the probability vector by ${\ell}_1$ normalization on the similarity
vector. $k$ is set to $5$ in our experiment. The parameter $\gamma$ is set to be $0.05$. We combine the probability vectors from different streams using a
simple weighted average rule.

\emph{Single-layer NN fusion.}
We first concatenate discriminative codes from IR and Flow
nets. Then, we construct a single softmax layer `shallow' neural network with
concatenated discriminative codes as inputs and action classes as outputs.

\emph{Two-layer NN fusion.}
We train a two-layer neural network consisting of one
$1\times 1$ convolution layer and a softmax output layer, with
concatenated discriminative codes from IR and Flow nets as
inputs. The convolution layer can play a role in selecting
good features from the input concatenated features. We used $50$ filters for the convolutional layer.

Table~\ref{tb3} presents the results of these fusion
strategies. Simple weighted rule for `Late fusion1' and `Late fusion2' can lead to performance improvement over single stream CNN. The more complex single-layer NN and two-layer NN fusion
methods do not outperform the simple weighted average fusion rule, likely due to a limited size of the InfAR dataset, since there is insufficient training samples to learn the parameters of convolution and softmax layers in these two fusion networks.

If we use a simple average rule to combine the confidence scores for all actions for each video from the method `Late fusion2' with the scores generated from the method `Early fusion of all concepts' in section~\ref{sec:exp:lowlevel}, we can achieve $79.2\%$ AP. This means that high-level concept features can provide complementary information to 3D CNN features for action recognition.

\begin{table}
\centering
\begin{tabular} {|c|c|}
\hline
Method & AP (\%)\\
\hline
Late fusion 1& 74 \\
Late fusion 2& 77.5\\
Single-layer NN fusion & 71.25\\
Two-layer NN fusion & 70.42\\
\hline
\end{tabular}
\vspace{5pt}
\caption{Recognition performances of fusion with 3D CNN features from IR and Flow nets. Please refer to the section `Fusion of two 3D CNNs' for the details of these fusion methods. }
\label{tb3}
\vspace{-0.3cm}
\end{table}

\begin{figure}
\begin{center}
\centering
\subfigure[IR stream]
{
\label{fig3-2:a}
\includegraphics[width=0.72\linewidth]{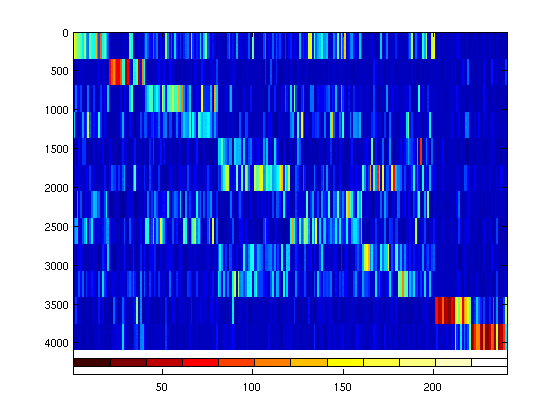}
}\\
\vspace{-0.4cm}
\subfigure[Flow stream]
{
\label{fig3-2:b}
\includegraphics[width=0.72\linewidth]{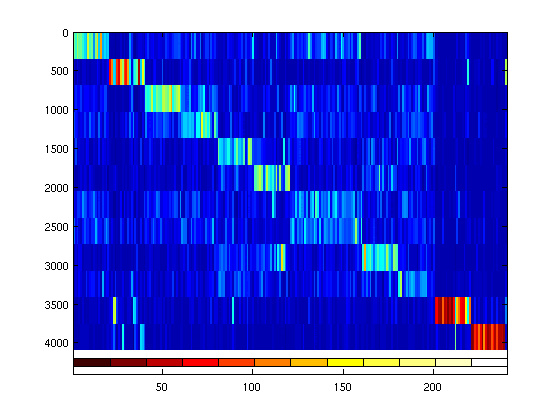}
}
\end{center}
\vspace{-0.4cm}
\caption{Visualization of learned discriminative codes of testing videos. Y axis indicates the dimensions of
  predicted codes while each position in X axis correspond to one test video. Blue color indicates `low value' while red color indicates `high value'.
  Each color in the color bar located at the bottom of each subfigure represents one action class for a subset of testing videos. }
\label{fig3-2}
\vspace{-0.5cm}
\end{figure}

\begin{figure}
\begin{center}
\centering
\subfigure[IR net]
{
\label{fig3-3:a}
\includegraphics[width=0.9\linewidth]{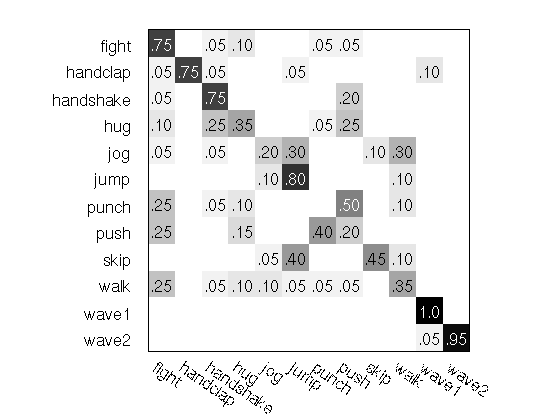}
}\\
\vspace{-0.4cm}
\subfigure[Flow net]
{
\label{fig3-3:b}
\includegraphics[width=0.9\linewidth]{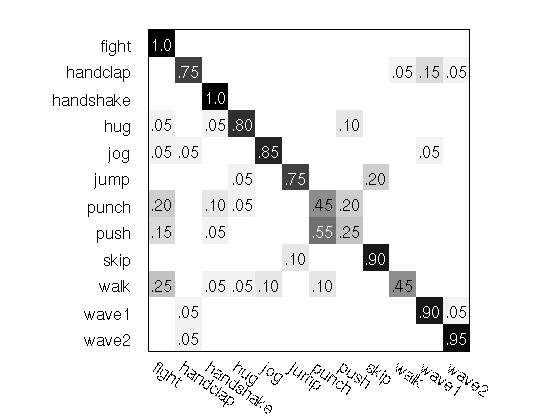}
}\\
\vspace{-0.4cm}
\subfigure[Late fusion of IR and flow nets]
{
\label{fig3-3:c}
\includegraphics[width=0.9\linewidth]{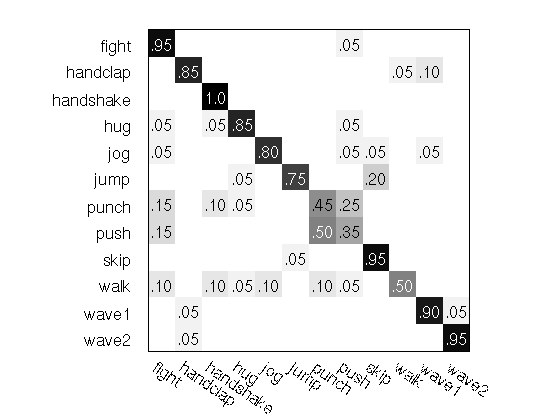}
}
\end{center}
\vspace{-0.4cm}
\caption{Confusion matrices for infrared action
  recognition. The 3D CNNs, `IR net' and `Flow net',
  are trained with softmax loss and
  discriminative code loss. Weighted average
  rule is used to fuse IR net's and Flow net's prediction scores.}
\label{fig3-3}
\vspace{-0.5cm}
\end{figure}

\subsection{Discussion}

Figure~\ref{fig3-2} visualizes the learned discriminative
codes of testing videos using both IR and Flow nets. The discriminative
predicted code matrix of all testing videos should be block-diagonal. Y
axis indicates the dimension of predicted code, and
each position in X axis correspond to one test video. As can
be seen from the figure, the videos from the same class have
similar representations, while videos from different classes
have different representations.

Figure~\ref{fig3-3} shows confusion matrices obtained using IR net,
flow net and late fusion of two nets. The misclassifications
are mainly related to `push' and `punch' categories which are visually
similar. `walk' is misclassified as 'fight', which is
possibly caused by presence of moving people in the background.
The fusion of two nets helps correcting some typical
misclassifications, such as the confusion between `handclap' and `wave1' classes.

For $5$ action categories, we achieved precisions higher than $90$\% when using $30$ positive video samples per category.
Figure~\ref{fig3-5} shows some samples from three of these classes.

\begin{figure}
\begin{center}
\centering
\subfigure[]
{
\label{fig3-4:a}
\includegraphics[width=0.75\linewidth]{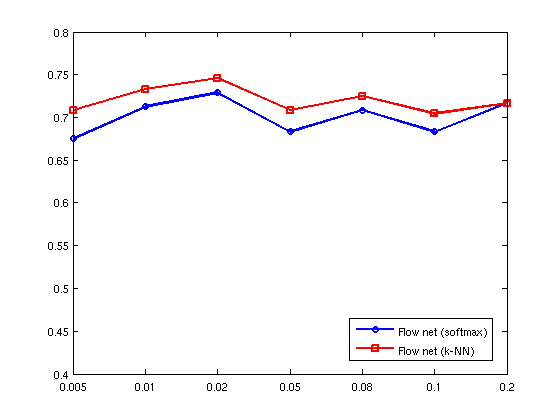}
}\\
\vspace{-0.4cm}
\subfigure[]
{
\label{fig3-4:b}
\includegraphics[width=0.75\linewidth]{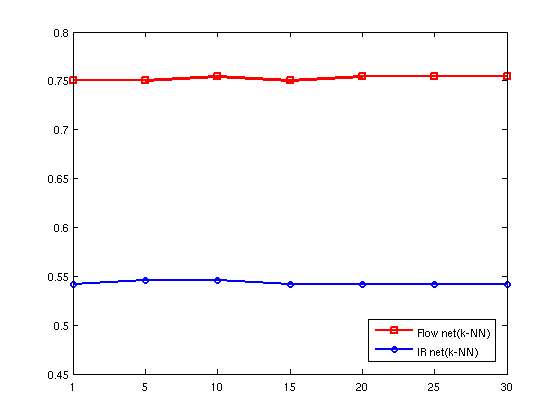}
}
\end{center}
\vspace{-0.4cm}
\caption{Effects of parameter selection of $\alpha$ and $k$ on the average precision performance on the InfAR dataset. (a)Effects of parameter selection of cost-balancing hyper-parameter $\alpha$. (b) Effects of  parameter  selection of $k$-NN neighborhood  size $k$.}
\label{fig3-4}
\vspace{-0.4cm}
\end{figure}

In Figure~\ref{fig3-4:a}, we plot the performance curves for a range of parameter $\alpha$ in flow net. We observe that our approach is not sensitive to the selection of $\alpha$. In addition, the simple $k$-NN classification scheme consistently outperforms the `softmax' classification on the full $\alpha$ range, this is because the generated predicted codes using our approach are discriminative. In Figure~\ref{fig3-4:b}, we show the performances using different $k$ (recall $k$ is the number of nearest neighbors for a $k$-NN classifier) for the IR and flow nets. Our approach is not sensitive to $k$ due to the increase of inter-class distances in the discriminative code space.

\begin{figure}
\begin{center}
\centering
\subfigure[Handshake, AP: $100\%$]
{
\label{fig3-5:a}
\includegraphics[width=0.95\linewidth]{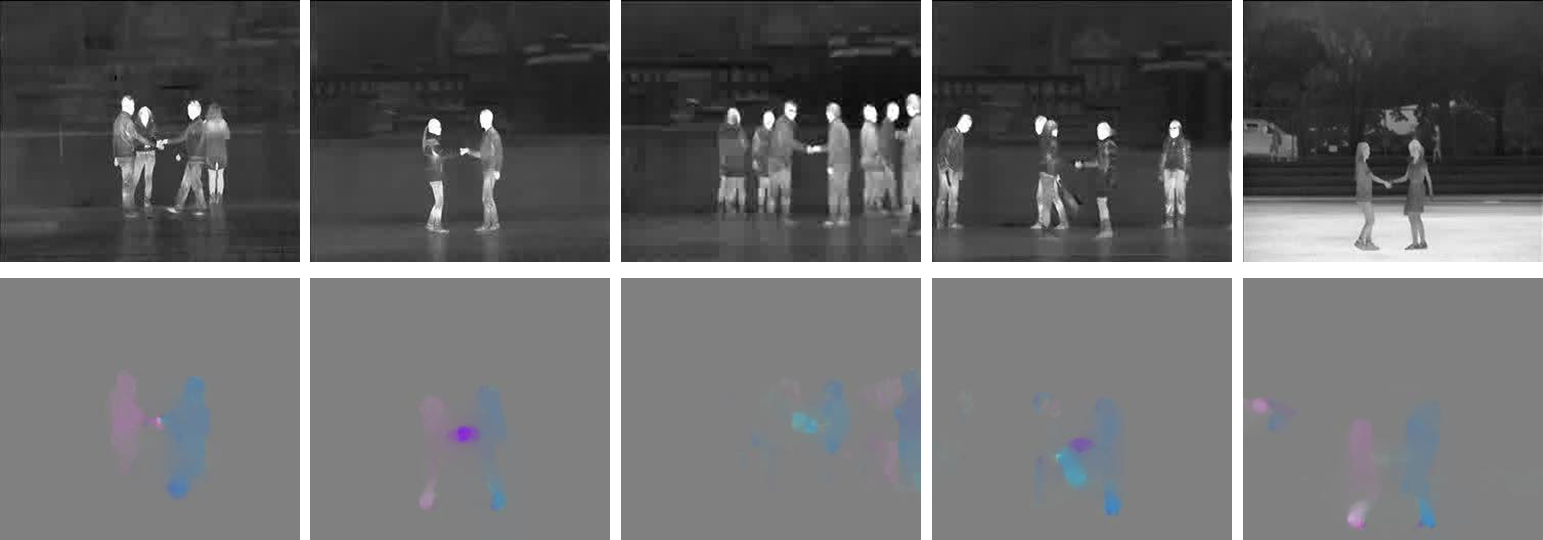}
}\\
\vspace{-0.2cm}
\subfigure[Fight, AP: $95\%$]
{
\label{fig3-5:b}
\includegraphics[width=0.95\linewidth]{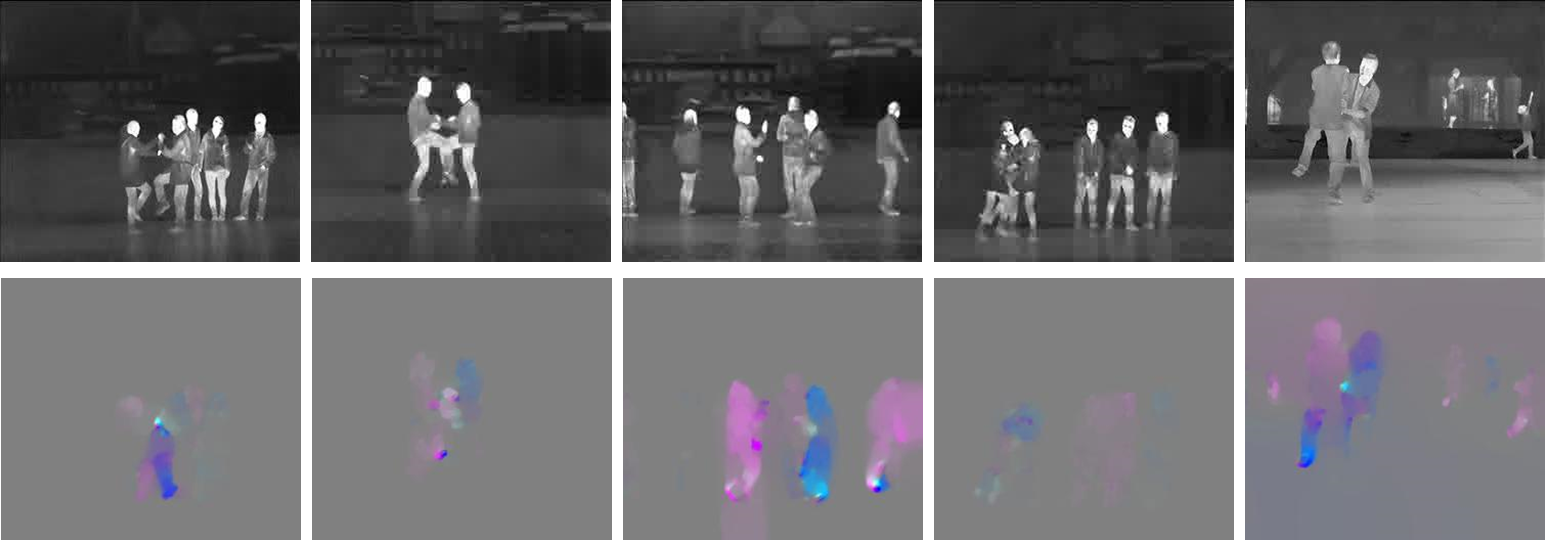}
}\\
\vspace{-0.2cm}
\subfigure[Wave2, AP: $95\%$]
{
\label{fig3-5:c}
\includegraphics[width=0.95\linewidth]{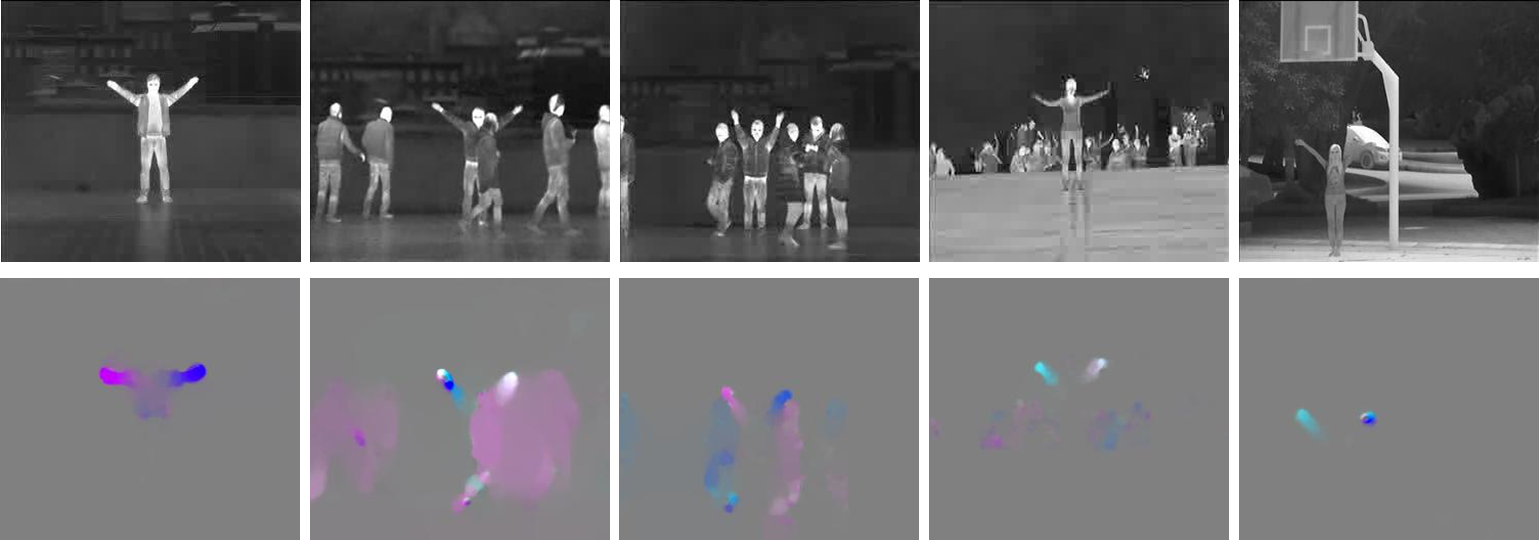}
}
\end{center}
\vspace{-0.3cm}
\caption{Video examples from classes with the high classification precision from the InfAR dataset. Each pair of images (IR and optical flow) is sampled from different testing videos and their corresponding optical flow fields.}
\label{fig3-5}
\vspace{-0.4cm}
\end{figure}

\section{Conclusion}

We introduce a two-stream 3D convolutional network for
action recognition in infrared videos. Each recognition
stream (IR and flow nets), was trained with softmax classification
loss and discriminative code loss making the extracted
representations of infrared videos become more discriminative. Both
nets are initialized by pretraining on visible spectrum videos, and finetuned on the
infrared videos. Our experiments show that using even a single flow stream CNN can achieve
state-of-the-art performance on the InfAR dataset. The goals of our
future work are to extend the current approach to the cross-spectral
feature learning and explore the domain adaptation techniques that can more effectively
exploit the high resource spectrum for the action recognition in the low-resource spectrum.

\section*{Acknowledgement}
This work is supported by the Intelligence Advanced Research Projects Activity (IARPA) via Department of Interior National Business Center contract number D11PC20071. The U.S. Government is authorized to reproduce and distribute reprints for Government purposes notwithstanding any copyright annotation thereon. Disclaimer: The views and conclusions contained herein are those of the authors and should not be interpreted as necessarily representing the official policies or endorsements, either expressed or implied, of IARPA, DoI/NBC, or the U.S. Government.

{\small
\bibliographystyle{ieee}
\bibliography{3DCNN}
}

\clearpage
\onecolumn
\section*{Appendix}

In order to explain why we can obtain discriminative predicted codes in Figure~\ref{fig3-2}, we visualize the learned features from the discriminative code layers in IR and flow nets via the gradient ascent approach~\cite{JYosinski15}. By starting at a $16$-frame sequence of randomly initialized images with dimension $16\times112\times112\times3$, this gradient ascent based approach can produce a sequence of optimized images (16 frame long) that cause high activation of the neuron in question. This is quite different to the \textit{deconv} visualization approach presented in ~\cite{Tran15}, which begins with an input video clip. Figure~\ref{fc7.5layer} visualizes these optimized images sequences for a subset of neurons in the discriminative code layer, which produces the $4096$-dimension output (\textit{i.e.}, predicted code) of the network. Please note that we follow~\cite{jiang17} and uniformly allocate neurons of this layer to $12$ action classes as described in Sec. 2.1. Here we only selected a subset of neurons which are assigned to a particular class during network training. Three neurons for each class from both IR and flow nets are visualized in this figure. Note that we did not do any fine-tuning of 3D CNN model on the InfAR dataset during this visualization step.

From this figure, we can know that these neurons can learn class-specific spatiotemporal patterns. For example, the `fight' neurons in IR net can learn body-parts and their shape in the starting and ending frames, and detect salient fight-like motion patterns in the other frames, which means that 3D CNN can model appearance and motion information simultaneously. The `fight' neurons in the flow net attend to salient motion patterns for action `fight'. In addition, these three neurons assigned to the same class can learn different discriminative spatiotemporal patterns, which consider multimodal distribution of signals of the same class.

\begin{figure*}[t]
\begin{center}
\includegraphics[width=1.0\linewidth]{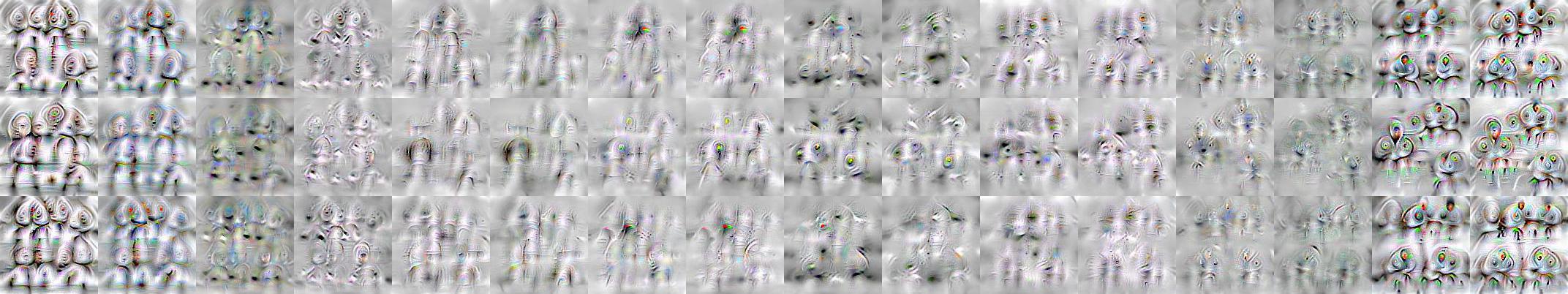} \\\vspace{0.05cm}
\includegraphics[width=1.0\linewidth]{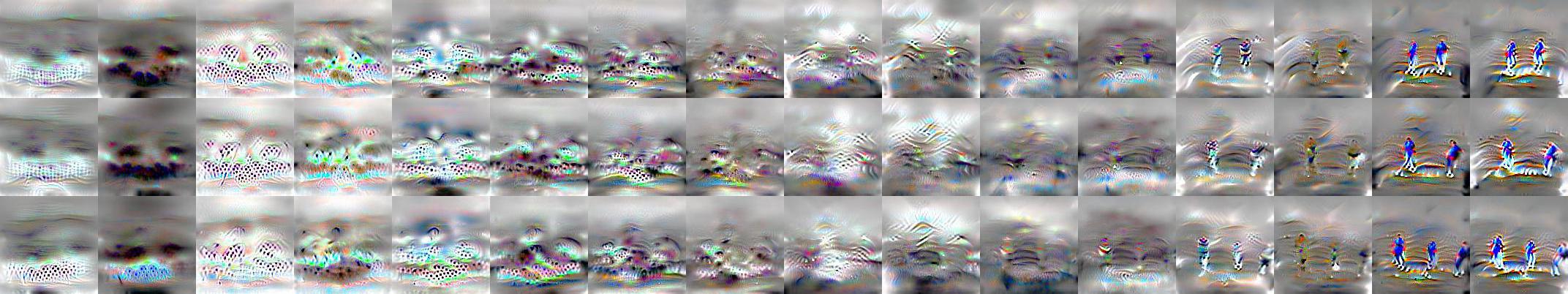} \\
(a) Neurons $0\sim2$, assigned to class `fight' (first three rows: IR net, other rows: flow net) \vspace{0.1cm} \\
\includegraphics[width=1.0\linewidth]{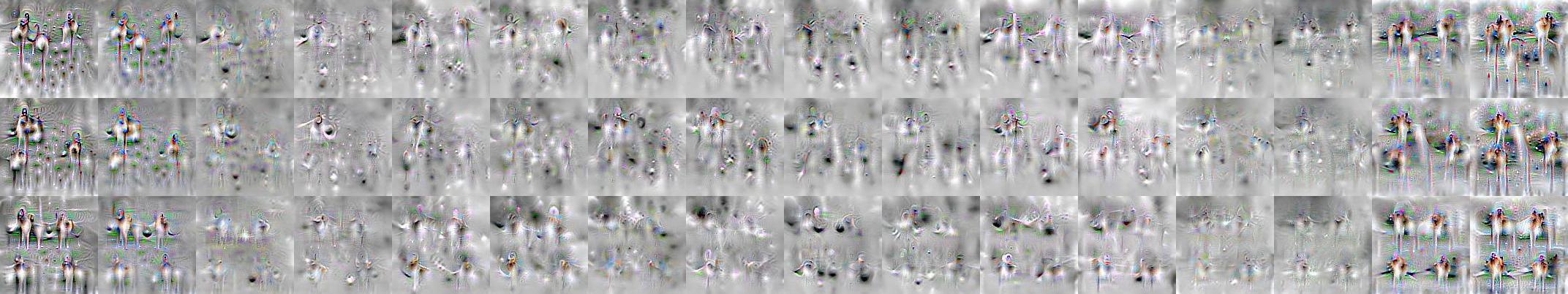} \\\vspace{0.05cm}
\includegraphics[width=1.0\linewidth]{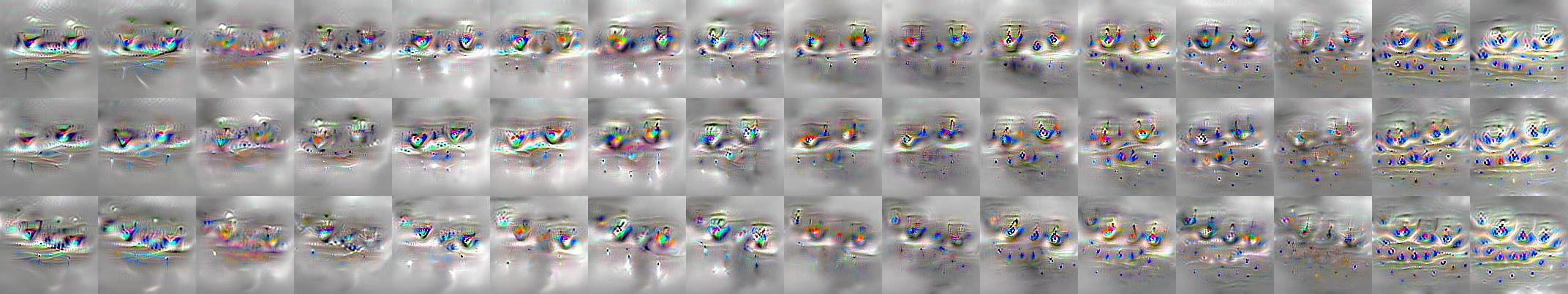} \\
(b) Neurons $341\sim343$, assigned to class `handclap'\vspace{0.1cm} \\
\includegraphics[width=1.0\linewidth]{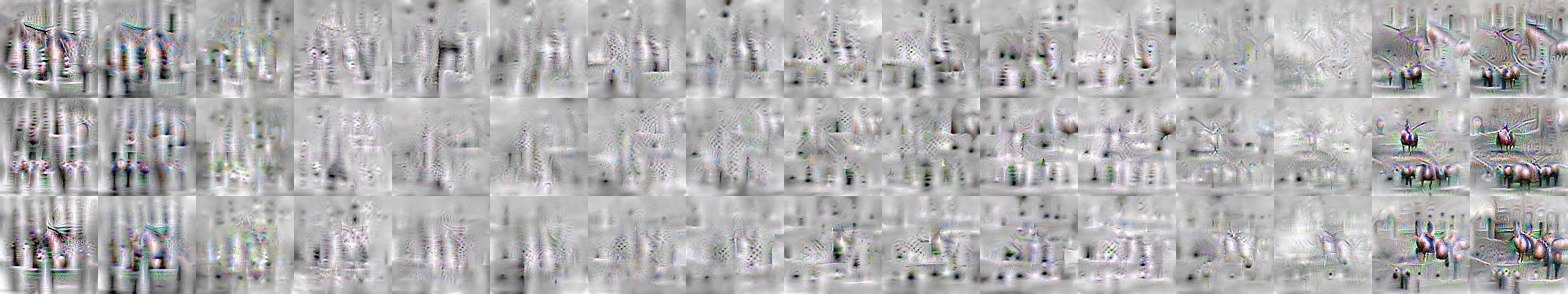} \\\vspace{0.05cm}
\includegraphics[width=1.0\linewidth]{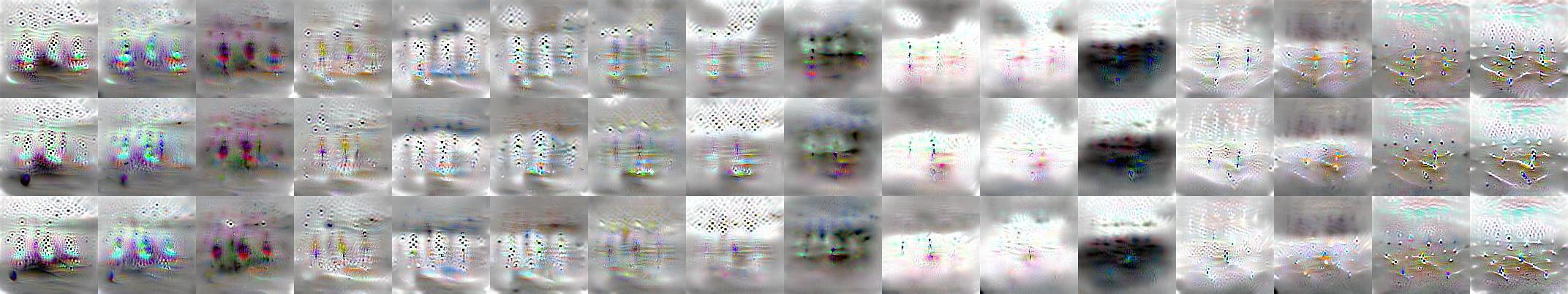} \\
(c) Neurons $682\sim684$, assigned to class `handshake' \vspace{0.1cm} \\
\end{center}
\end{figure*}
\begin{figure*}
\begin{center}
\includegraphics[width=1.0\linewidth]{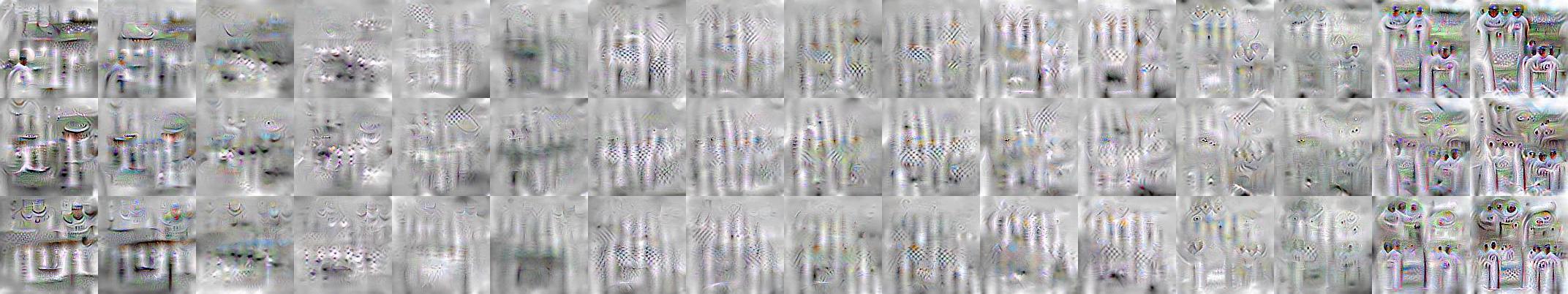} \\\vspace{0.05cm}
\includegraphics[width=1.0\linewidth]{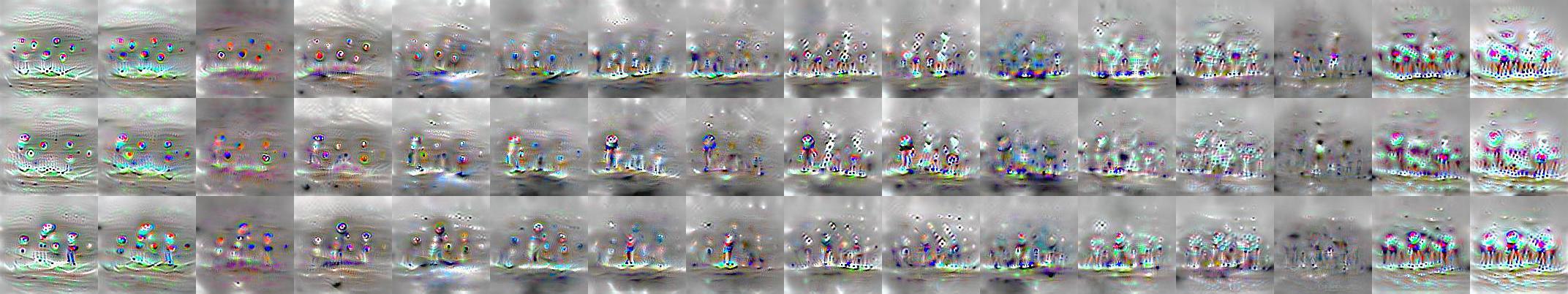} \\
(d) Neurons $1023\sim1025$, assigned to class `hug' \vspace{0.2cm} \\
\includegraphics[width=1.0\linewidth]{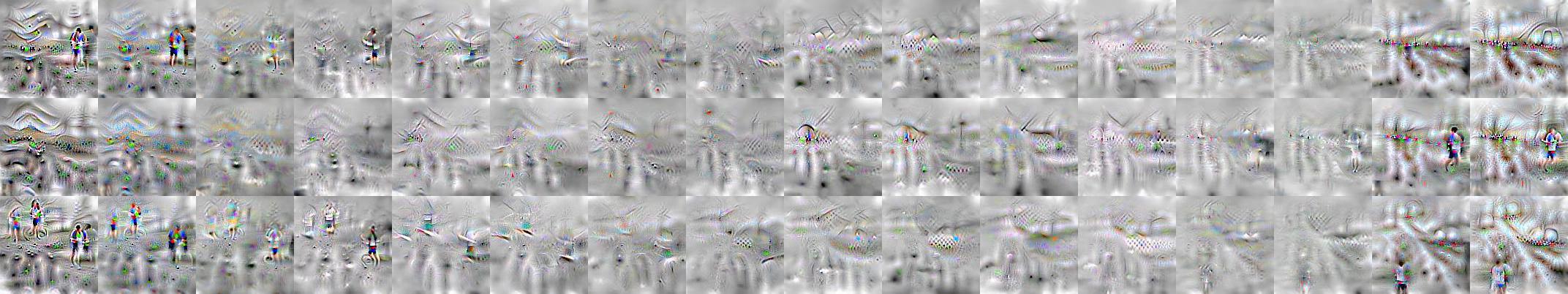} \\\vspace{0.05cm}
\includegraphics[width=1.0\linewidth]{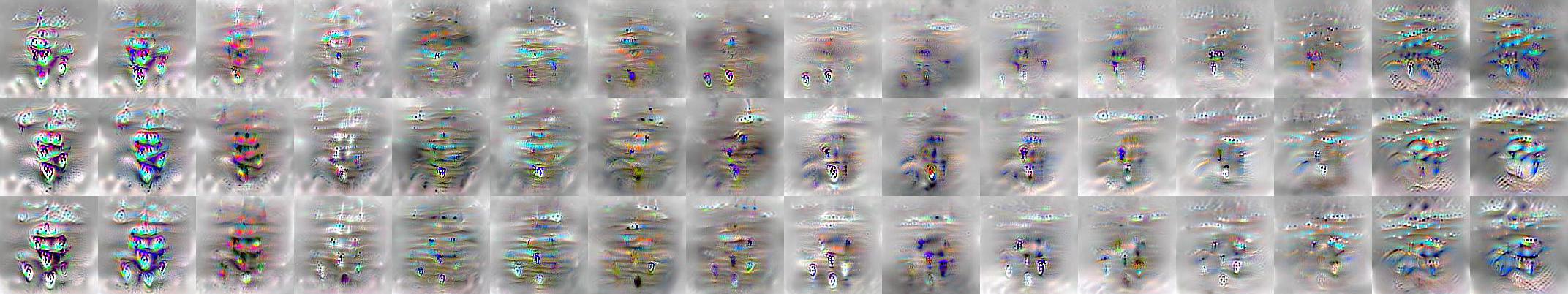} \\
(e) Neurons $1364\sim1366$, assigned to class `jog' \vspace{0.2cm} \\
\includegraphics[width=1.0\linewidth]{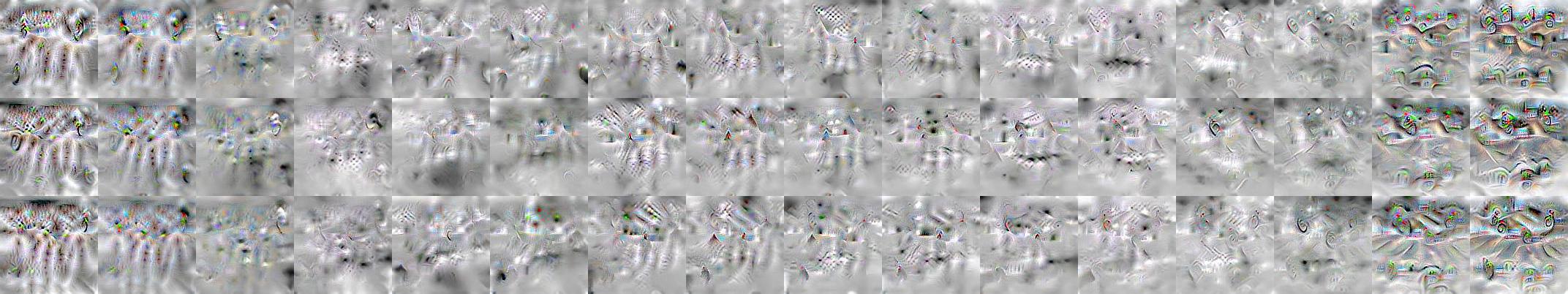} \\\vspace{0.05cm}
\includegraphics[width=1.0\linewidth]{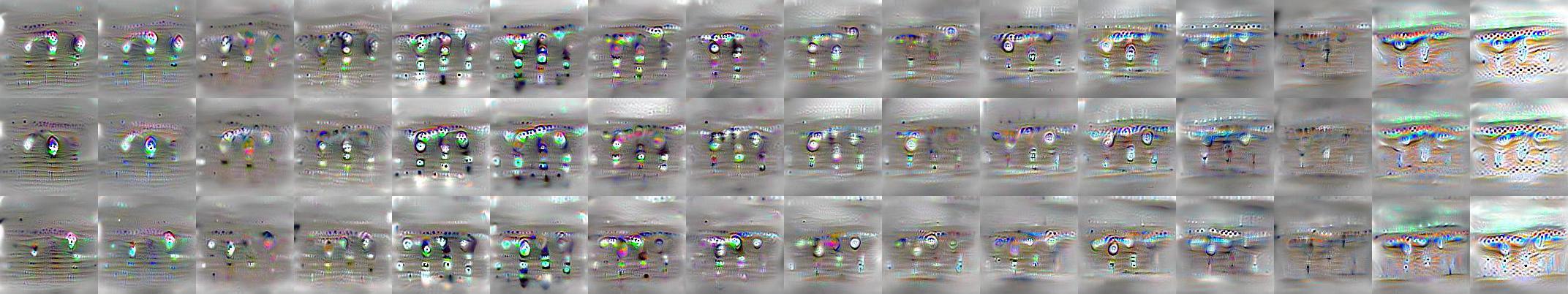} \\
(f) Neurons $1705\sim1707$, assigned to class `jump' \vspace{0.2cm} \\
\end{center}
\end{figure*}
\begin{figure*}
\begin{center}
\includegraphics[width=1.0\linewidth]{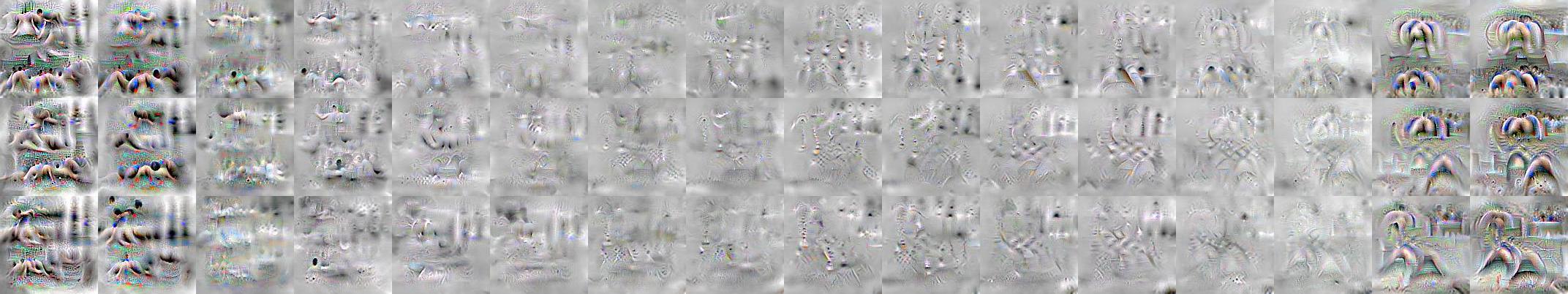} \\\vspace{0.05cm}
\includegraphics[width=1.0\linewidth]{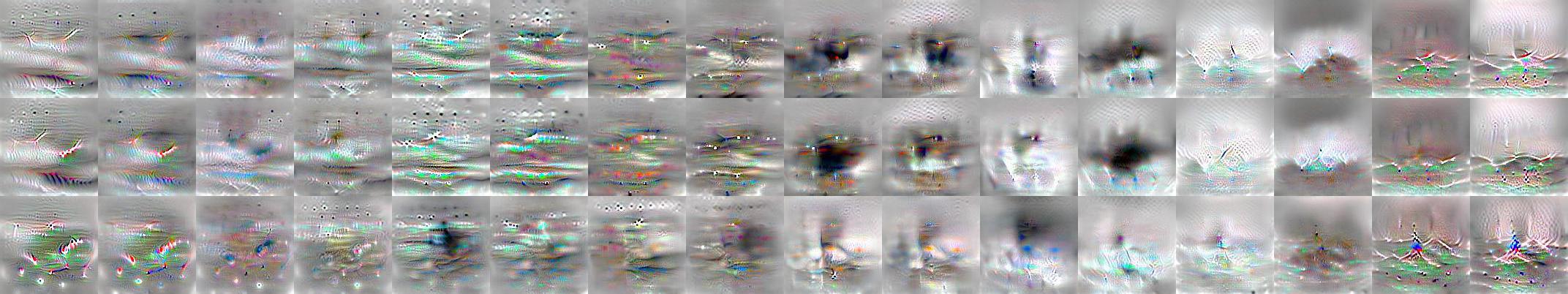} \\
(g) Neurons $2046\sim2048$, assigned to class `punch' \vspace{0.2cm} \\
\includegraphics[width=1.0\linewidth]{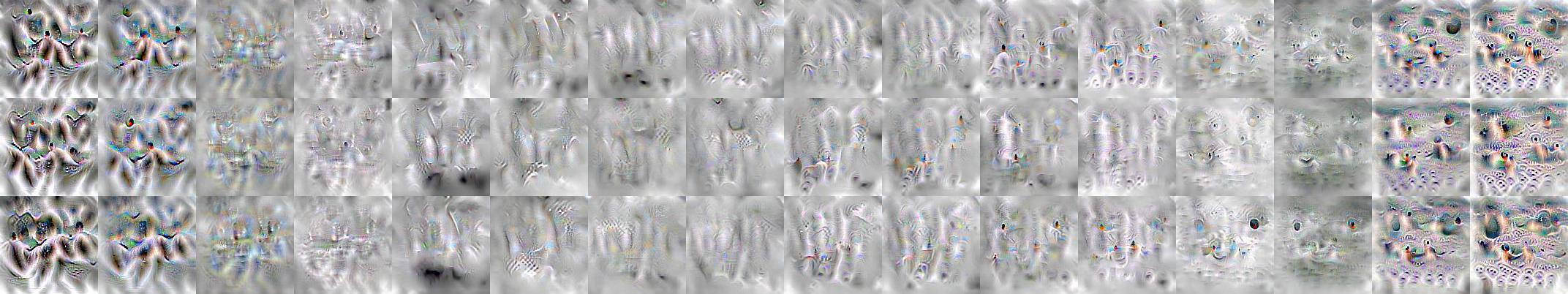} \\\vspace{0.05cm}
\includegraphics[width=1.0\linewidth]{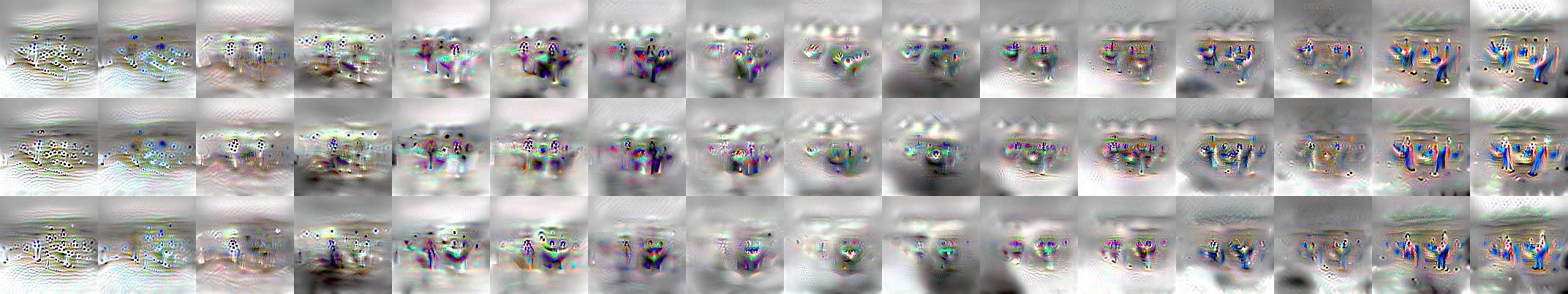} \\
(h) Neurons $2387\sim2389$, assigned to class `push' \vspace{0.2cm} \\
\includegraphics[width=1.0\linewidth]{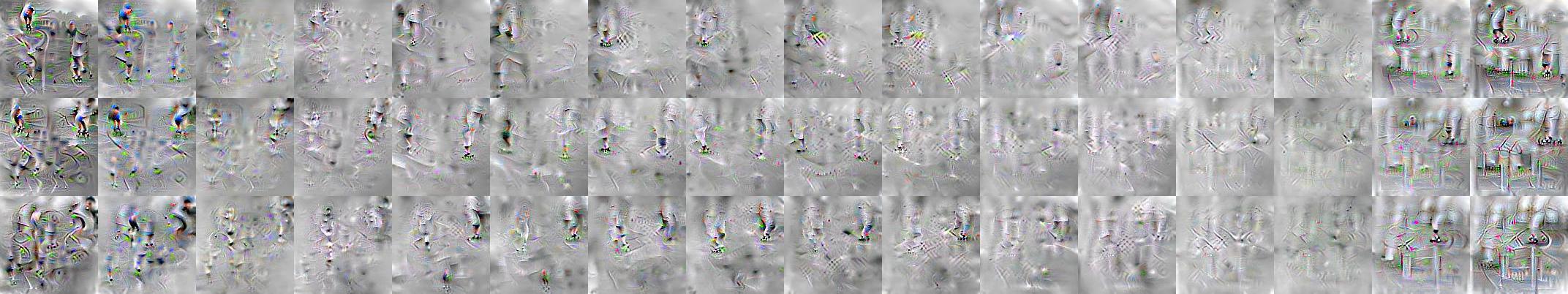} \\\vspace{0.05cm}
\includegraphics[width=1.0\linewidth]{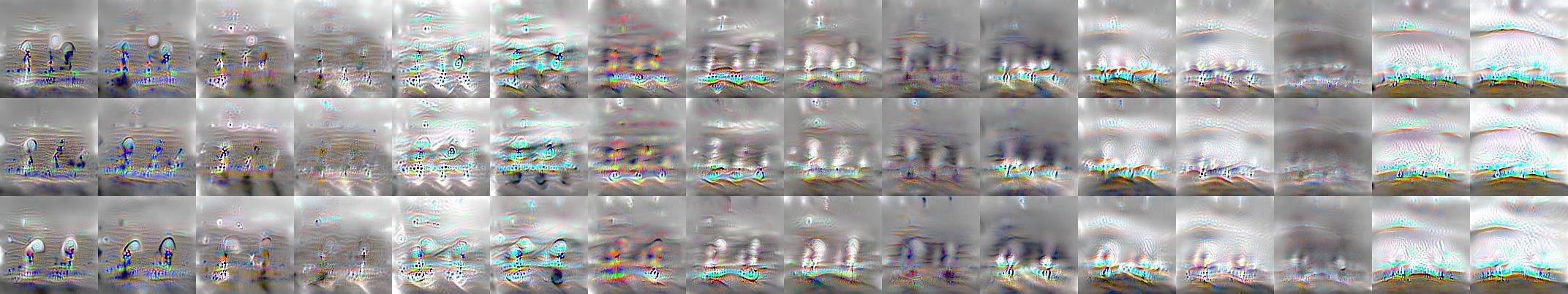} \\
(i) Neurons $2728\sim2730$, assigned to class `skip' \vspace{0.2cm} \\
\end{center}
\end{figure*}
\begin{figure*}
\begin{center}
\includegraphics[width=1.0\linewidth]{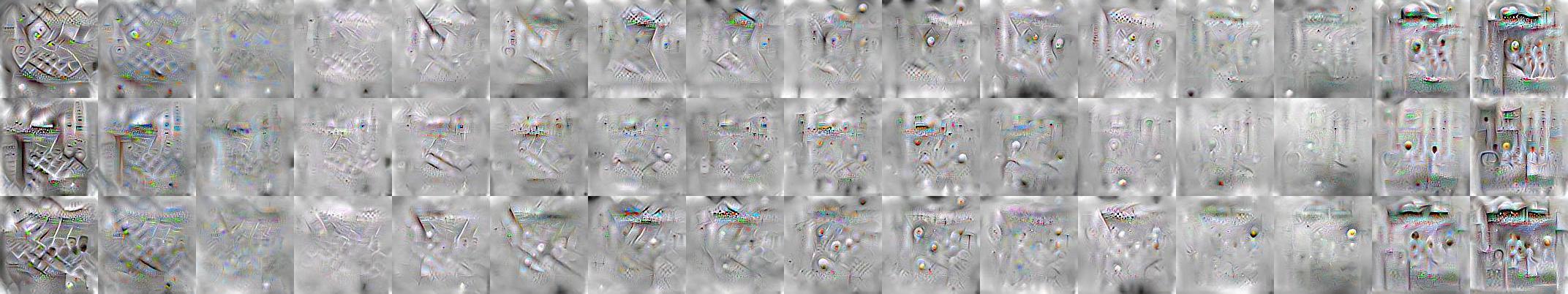} \\\vspace{0.05cm}
\includegraphics[width=1.0\linewidth]{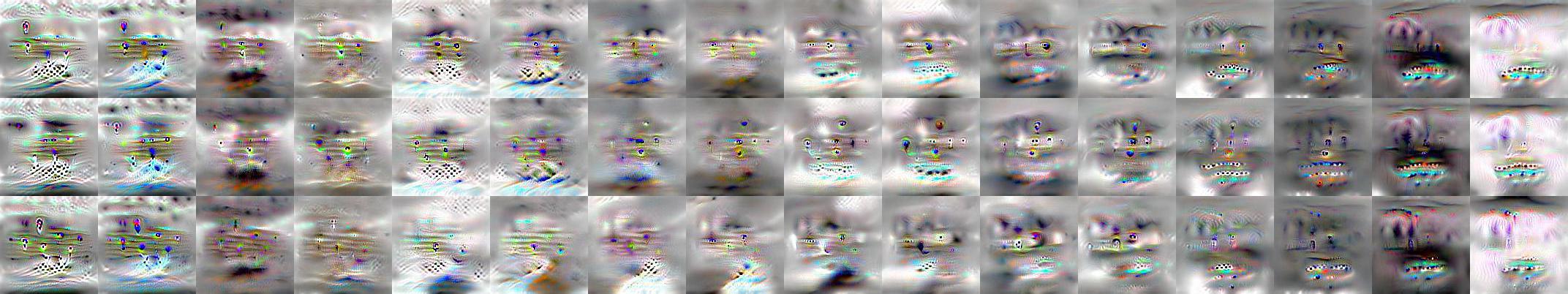} \\
(j) Neurons $3070\sim3072$, assigned to class `walk' \vspace{0.2cm}\\
\includegraphics[width=1.0\linewidth]{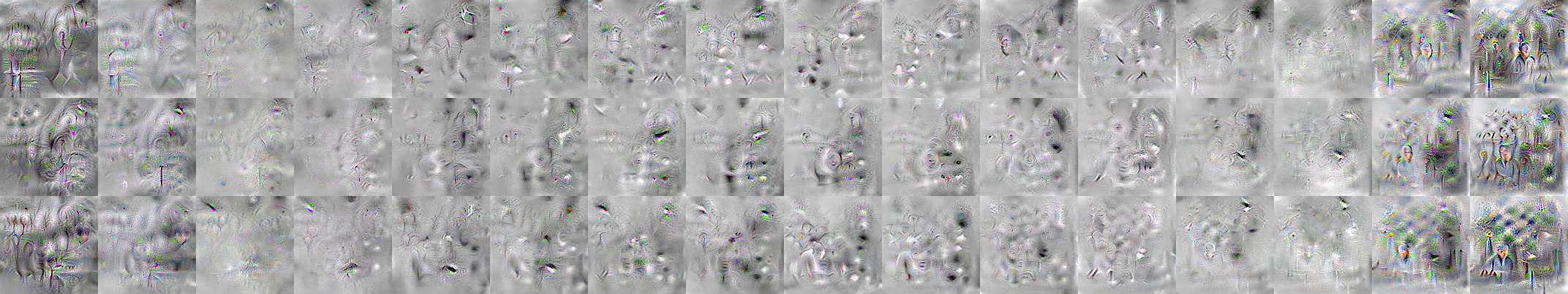} \\\vspace{0.05cm}
\includegraphics[width=1.0\linewidth]{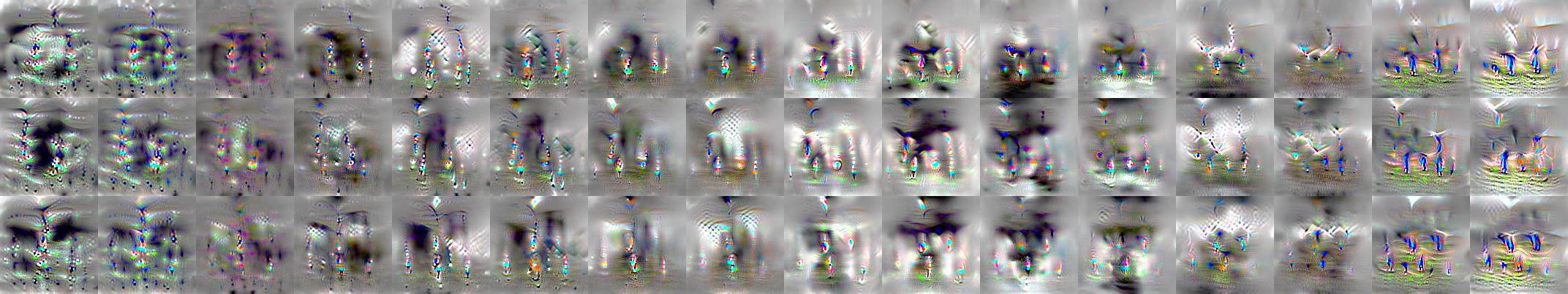} \\
(k) Neurons $3412\sim3414$, assigned to class `wave1' \vspace{0.2cm}\\
\includegraphics[width=1.0\linewidth]{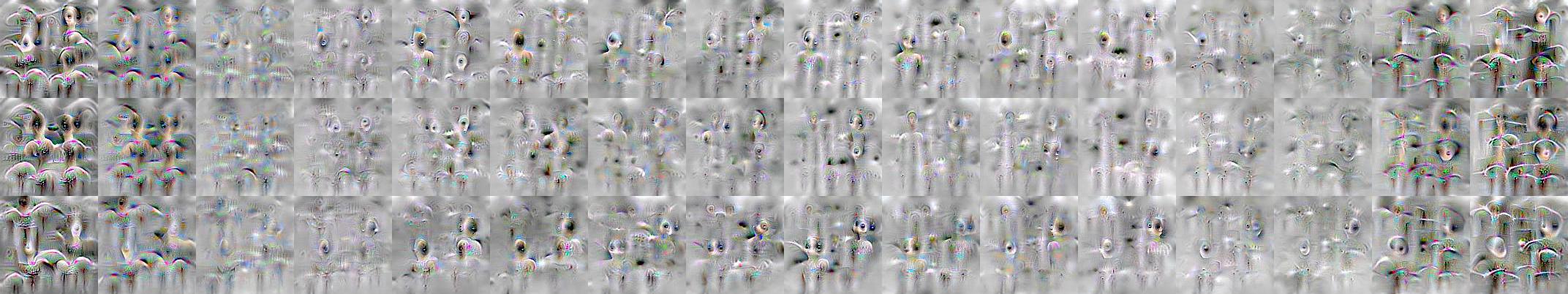} \\\vspace{0.05cm}
\includegraphics[width=1.0\linewidth]{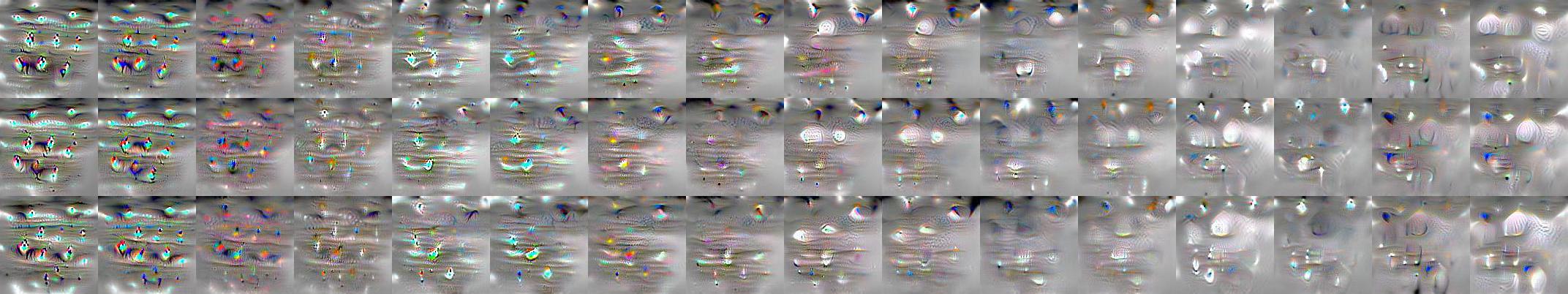} \\
(l) Neurons $3754\sim3756$, assigned to class `wave2' \vspace{0.2cm}\\
\end{center}
\caption{Visualization of learned class-specific neurons from the discriminative code layers in the IR and flow nets. In each subfigure, the first three rows visualize three neurons from IR net, while the other rows visualize the neurons from the flow net. These neurons can learn class-specific spatiotemporal features. The figure is best viewed in color and $400\%$ zoom in.}
\label{fc7.5layer}
\end{figure*}

\end{document}